\documentclass[conference]{IEEEtran}
\usepackage{blindtext, graphicx}

\usepackage{hyperref}

\usepackage{multicol}

\usepackage{multirow}

\usepackage{tabulary}

\usepackage[
backend=biber,
style=ieee,
sorting=nyt
]{biblatex}

\ifCLASSINFOpdf
\else
\fi

\begin{document}
%
% paper title
% can use linebreaks \\ within to get better formatting as desired
\title{A Real-Time Autonomous Highway Accident Detection Model Based on Big Data Processing and Computational Intelligence}

% author names and affiliations
% use a multiple column layout for up to three different
% affiliations
\author{
\IEEEauthorblockN{Murat Ozbayoglu }
\IEEEauthorblockA{Computer Engineering Department\\
TOBB Univ. of Economics and Tech.\\
Ankara, Turkey\\
Email: mozbayoglu@etu.edu.tr}
\and
\IEEEauthorblockN{Gokhan Kucukayan}
\IEEEauthorblockA{Computer Engineering Department\\
TOBB Univ. of Economics and Tech.\\
Ankara, Turkey\\
Email: gokhankucukayan@gmail.com}
\and
\IEEEauthorblockN{Erdogan Dogdu}
\IEEEauthorblockA{Computer Engineering Department\\
TOBB Univ. of Economics and Tech.\\
Ankara, Turkey\\
Email: erdogandogdu@gmail.com}
}

% conference papers do not typically use \thanks and this command
% is locked out in conference mode. If really needed, such as for
% the acknowledgment of grants, issue a \IEEEoverridecommandlockouts
% after \documentclass

% for over three affiliations, or if they all won't fit within the width
% of the page, use this alternative format:
% 
%\author{\IEEEauthorblockN{Michael Shell\IEEEauthorrefmark{1},
%Homer Simpson\IEEEauthorrefmark{2},
%James Kirk\IEEEauthorrefmark{3}, 
%Montgomery Scott\IEEEauthorrefmark{3} and
%Eldon Tyrell\IEEEauthorrefmark{4}}
%\IEEEauthorblockA{\IEEEauthorrefmark{1}School of Electrical and Computer Engineering\\
%Georgia Institute of Technology,
%Atlanta, Georgia 30332--0250\\ Email: see http://www.michaelshell.org/contact.html}
%\IEEEauthorblockA{\IEEEauthorrefmark{2}Twentieth Century Fox, Springfield, USA\\
%Email: homer@thesimpsons.com}
%\IEEEauthorblockA{\IEEEauthorrefmark{3}Starfleet Academy, San Francisco, California 96678-2391\\
%Telephone: (800) 555--1212, Fax: (888) 555--1212}
%\IEEEauthorblockA{\IEEEauthorrefmark{4}Tyrell Inc., 123 Replicant Street, Los Angeles, California 90210--4321}}

% use for special paper notices
%\IEEEspecialpapernotice{(Invited Paper)}

% make the title area
\maketitle

\begin{abstract}
%\boldmath

Due to increasing urban population and growing number of motor vehicles, traffic congestion is becoming a major problem of the 21st century. One of the main reasons behind traffic congestion is accidents which can not only result in casualties and losses for the participants, but also in wasted and lost time for the others that are stuck behind the wheels. Early detection of an accident can save lives, provides quicker road openings, hence decreases wasted time and resources, and increases efficiency. In this study, we propose a preliminary real-time autonomous accident-detection system based on computational intelligence techniques. Istanbul City traffic-flow data for the year 2015 from various sensor locations are populated using big data processing methodologies. The extracted features are then fed into a nearest neighbor model, a regression tree, and a feed-forward neural network model. For the output, the possibility of an occurrence of an accident is predicted. The results indicate that even though the number of false alarms dominates the real accident cases, the system can still provide useful information that can be used for status verification and early reaction to possible accidents.

\end{abstract}
% IEEEtran.cls defaults to using nonbold math in the Abstract.
% This preserves the distinction between vectors and scalars. However,
% if the journal you are submitting to favors bold math in the abstract,
% then you can use LaTeX's standard command \boldmath at the very start
% of the abstract to achieve this. Many IEEE journals frown on math
% in the abstract anyway.

% Note that keywords are not normally used for peerreview papers.
\begin{IEEEkeywords}
Traffic flow, big data, accident detection, intelligent transportation systems, neural networks, nearest neighbor, regression tree, computational intelligence, machine learning, IoT, sensors.
\end{IEEEkeywords}

% For peer review papers, you can put extra information on the cover
% page as needed:
% \ifCLASSOPTIONpeerreview
% \begin{center} \bfseries EDICS Category: 3-BBND \end{center}
% \fi
%
% For peerreview papers, this IEEEtran command inserts a page break and
% creates the second title. It will be ignored for other modes.
\IEEEpeerreviewmaketitle

\section{Introduction}

Disruption of normal traffic flow results in wasted time, higher fuel costs and lost productivity. Officials are working around the clock to finish the regular maintenance activities as quickly as possible to keep the roads open and provide safe traffic flow. Meanwhile, traffic accidents are among the most important causes that disrupt the normal traffic flow. 

Preventing an accident is important, however it is very difficult, if not impossible to provide an accident-free road vehicle transportation system. Even though it may not be possible to avoid accidents altogether, early detection of and reaction to accidents are very important in saving lives and reducing accident-related costs. This study aims to provide such a system where real-time traffic-flow data is monitored and the occurrence of accidents is predicted before any official accident notification arrives from the scene. This can for example help city governments to dispatch emergency teams to prospective accident-prone areas, prevent accidents by taking extra-measures, manage costs efficiently, and be well prepared for future events in general.

After this brief introduction, the rest of the paper is as follows. In Section 2, the related studies regarding traffic flow, traffic accident detection using different approaches, methodologies are provided. The data, domain and big data processing techniques used in this study are explained in Section 3 followed by Section 4 that gives the data analysis and feature extraction details. The results and the discussions are provided in Section 5. We conclude and provide open problems and future direction in Section 6.

\section{Previous Work}

There are a lot of studies about traffic and transportation systems; however most of them are concentrated on infrastructure development, enhancements on physical infrastructures, etc. Meanwhile some studies focus on some particular intelligent transportation system components such as the topics of accident prevention, traffic flow estimation, event detection, route optimization, etc. These aforementioned studies are mostly of our concern.

In one study, Lee concentrated on analyzing traffic data quality collected through various road sensors and observed the difficulties of obtaining error-free, reliable data~\cite{lee2012}. He provided methods for detecting some of these errors and tested these models using real data. Wang suggested using Road Traffic Microwave Sensors (RTMS) for more reliable traffic data for Traffic Management Systems and covered some examples that are being used in Ontario~\cite{wang1992road}. 

Masters et al~ \cite{masters1991incident} analyzed an early traffic event detection system called COMPASS, implemented in Toronto area, which was based on algorithms using density, road occupancy and velocity aiming to detect anomalies from the data.  Feng et al~\cite{feng2014probe} used similar data to calculate the average trip time using Bayes model. Pascale et al~\cite{pascale2015characterization} concentrated on road capacity calculations and suggested methods to increase efficiency. 

There are also some studies based on image and video processing techniques that rely on data obtained through CCTV cameras~\cite{jain2012road,buyukozcu2012}. The advantage of these systems is the possibility of observing different types of vehicles and their behaviors.

Vehicle Ad Hoc Networks are also used in traffic data collection models~\cite{kocckan2008tacsitlar,zeng2016opportunistic,yuan2014traffic}, but maintaining the communication and data feed is not easy under different road conditions.

Researchers also study traffic flow and event/accident detection. Baiocchi et al~\cite{baiocchi2015vehicular} suggested a traffic flow estimation system using GPS data and obtained successful results. In a similar study Terroso-Saenz et al~\cite{terroso2012cooperative} estimated the flow density. 

Hojati et al~\cite{hojati2014modelling} provided a model for estimating the surpassed time interval between the occurrence of an accident and clearing it depending on different road conditions and infrastructure differences.   
Computational intelligence techniques have also been part of some of the proposed systems for traffic flow estimation, accident detection, etc. Neural networks~\cite{lu2012hybrid,yildirim2015ankara,ccetiner2010neural}, Support Vector Machines~\cite{lu2012automatic,vsingliar2007learning} and Hidden Markov Model~\cite{akoz2014traffic} are used in these studies. 

In this study, we propose a model for accident detection using three separate computational intelligence techniques. Moreover, the model inputs are obtained using big data processing techniques in order to have scalability and real-time implementation capability. The main motivation for this approach is to be able to provide a real-time traffic monitoring system that can detect accidents immediately after the incident in order to take the necessary precautions as early as possible.

\section{Data Processing}

\subsection{Traffic Data}

\begin{figure*}
    \centering
    \includegraphics[width=\textwidth]{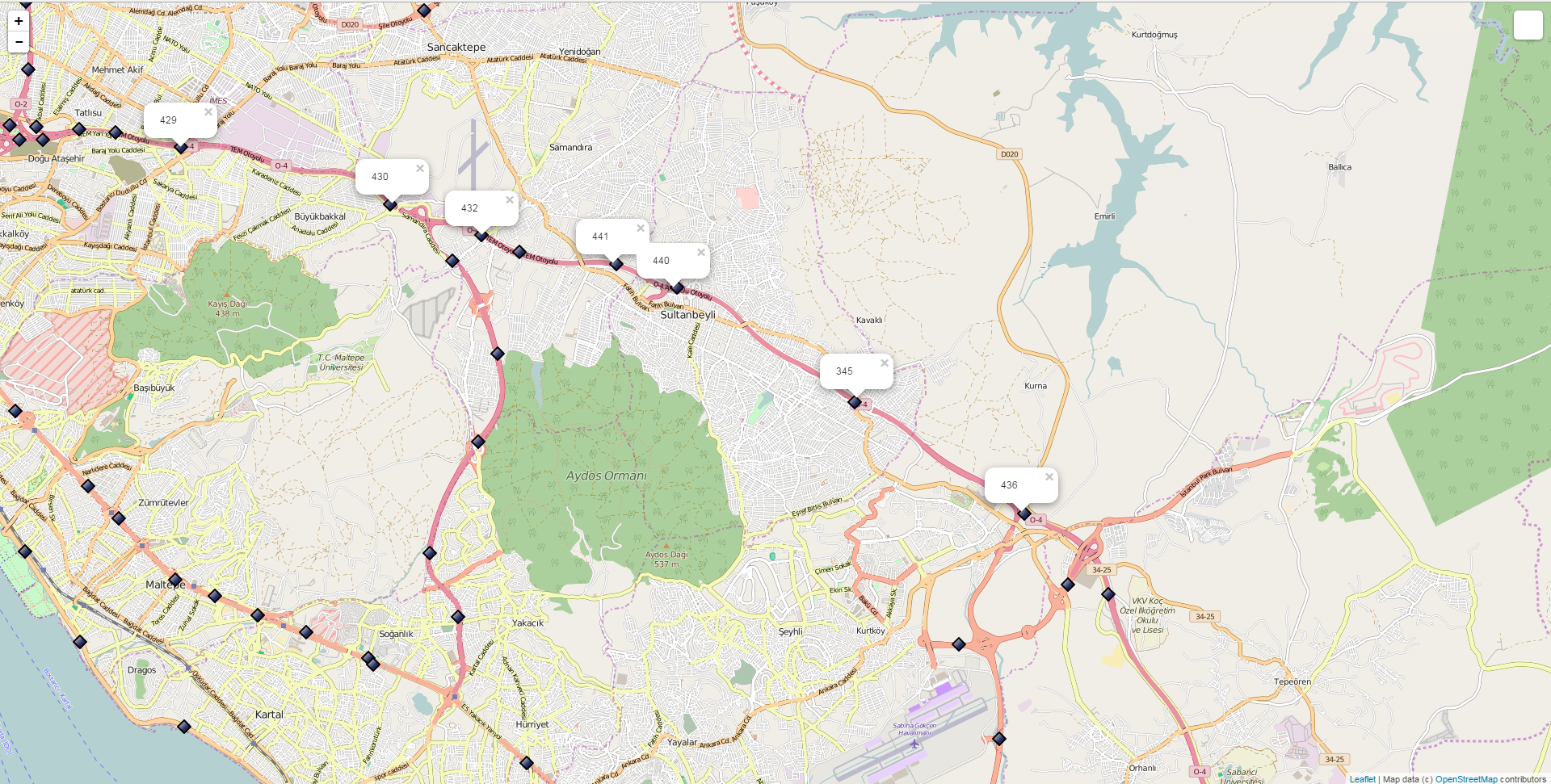}
    \caption{Sensor locations on the map.}
    \label{fig:map}
\end{figure*}

The data used in this study is provided by Istanbul Municipal Traffic Control Department\footnote{http://isbak.istanbul/intelligent-transportation-systems/}. The dataset consists of 2015 calendar year traffic flow data obtained from 7 separate RTMS (Real-Time Monitoring System) sensors on one of the major highways (TEM-Trans European Motorway). The location of the sensors can be seen in Figure~\ref{fig:map}. The distance between the first and the last sensor is almost 15 miles. 

There are approximately 400 sensors that feed data throughout the city area, however to implement the preliminary version of our model, we chose to isolate the data from these aforementioned sensors. The main reason for this particular choice is due to the fact that there are no traffic lights, stop signs, sharp curves, etc. on that section of the highway. Hence, our assumption was if any slowdown or stoppage is seen in the traffic, it should be due to a disruption on the road (an accident, road work, handicapped car, etc.).

The following data items are collected every 2 minutes through the sensors from each lane separately: 
\begin{itemize}
    \item Number of cars passing every 2 minutes.
    \item Average speed of the vehicles in the last 2 minutes.
    \item Average occupancy ratio of the lane in the last 2 minutes.
    \item Date/Time information.
\end{itemize}

\begin{figure*}
    \centering
    \includegraphics[width=\textwidth]{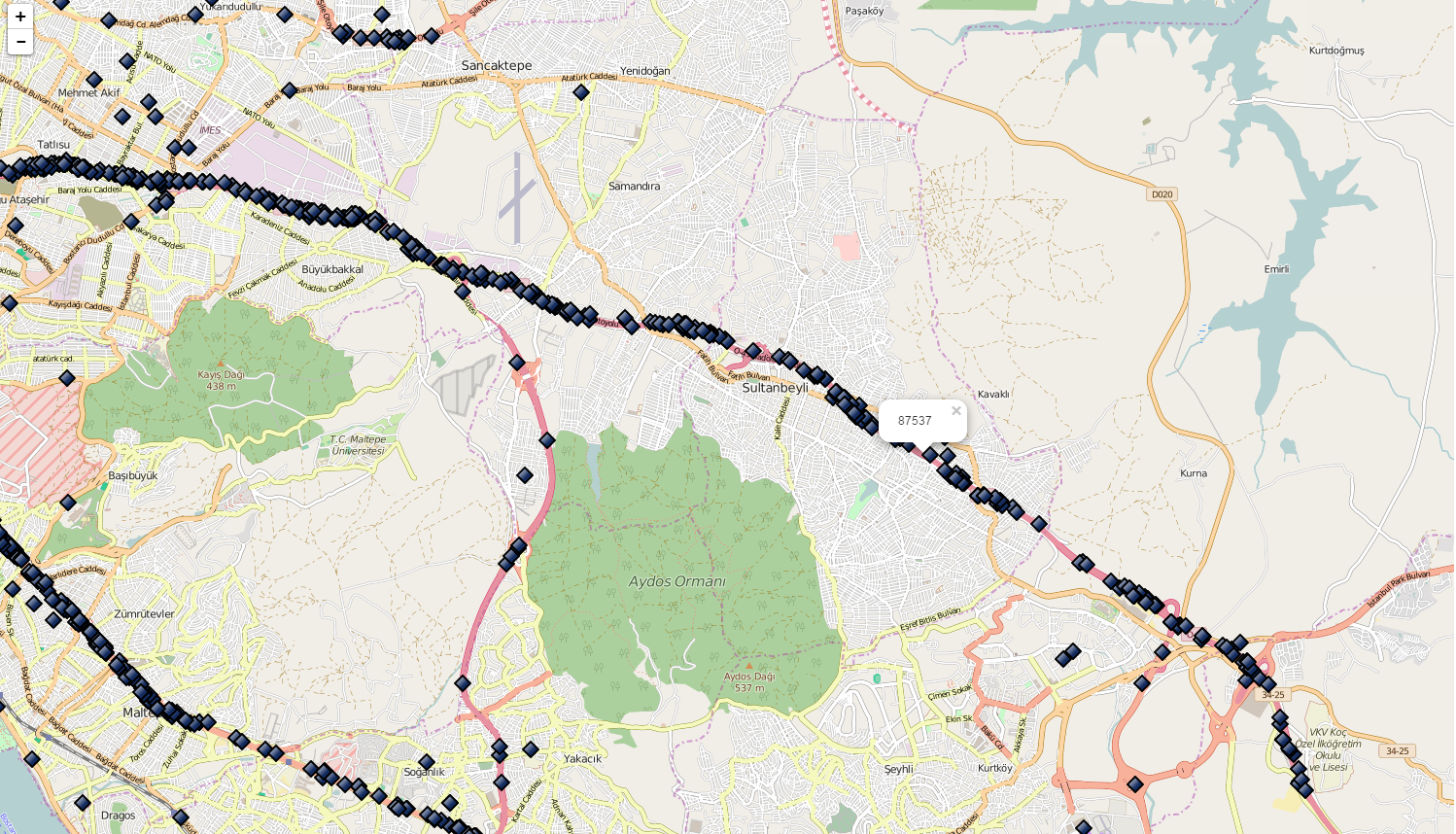}
    \caption{Incidents on the map.}
    \label{fig:map2}
\end{figure*}

Also, from the Traffic Department Database, the information about the accidents/disruptions, etc. are extracted. In the calendar year 2015, 72 incidents (events) were observed in that particular section of the highway. Figure~\ref{fig:map2} shows all instances in the neighborhood including the 72 incidents that are used in our study.

The event database has the date/time, the location and the basic description of the incidents including the direction and the type of incident (serious accident, minor accident, road work, handicapped car, etc.). 

\subsection{Big Data Processing}

The raw data acquired from the Traffic Department is not usable in its existing format due to the amount of the data and some inconsistencies in the syntax and content.  The total amount of collected information for one year from all sensors consists of more than 100 million rows of structured and unstructured data. 

Meanwhile, a lot of preprocessing on the raw data was necessary in order to standardize the featured data. Some of the major problems that were faced during the preprocessing are due to the fact that each sensor sends its own version of data, the number of incoming and outgoing lanes were not known, etc. In addition, some of the lane data is non-existent, missing or showing 0 or NULL value from time to time. For example, there might be lane closures, the sensors might be obstructed by a leaf, paper, or the sensor might be malfunctioning. 

\begin{figure}[h]
    \centering
    \includegraphics[width=2.5in]{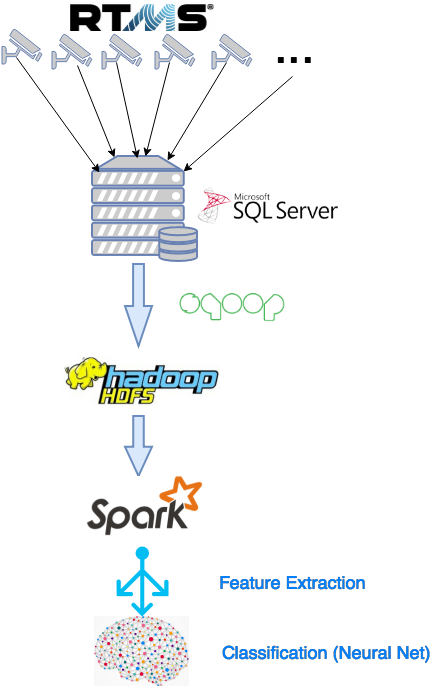}
    \caption{The flow of data.}
    \label{fig:process}
\end{figure}

In order to process this type of problematic data and extract the features accordingly, big data processing techniques are adapted. The acquired raw data is passed through an ETL (Extract-Transform-Load) process through HDFS and Apache Spark. The original data is stored in SQLServer format and imported to Hadoop environment via Sqoop (Figure~\ref{fig:process}).

The imported data is then processed on a 10-PC cluster using Spark and HDFS. The individual lane data are averaged in order to get a single directional feature. Also the inconsistencies within the data are eliminated accordingly. 

\section{Data Analysis And Feature Extraction}

\begin{figure*}[t]
    \centering
    \includegraphics[width=4in]{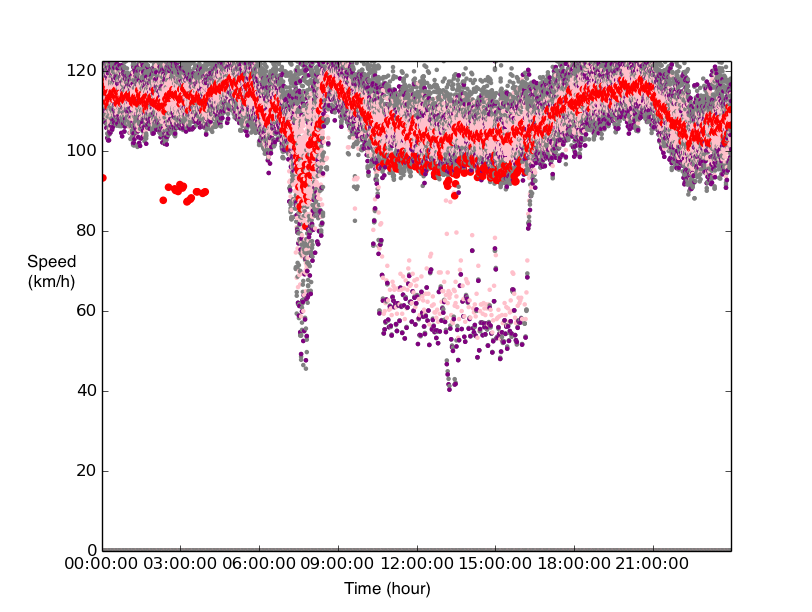}
    \caption{Sensor data.}
    \label{fig:sensdata}
\end{figure*}

The processed data is further analyzed to utilize the daily routines, seasonal behaviors and weekday/weekend effects. Even though this particular section of the highway is not close to downtown or city center, the traffic is still congested most of the time during rush hour on weekdays. The traffic flow pattern on one of these sensors observed only for Mondays for one direction is shown in Figure~\ref{fig:sensdata}.

In Figure~\ref{fig:sensdata}, it can be seen that average velocity is around 110 km/h, but during rush hours (especially in the morning), it slows down considerably. Evening rush hour is not noticeable for this particular sensor, this indicates the traffic direction is mostly into the city in the mornings, but not in the evenings.

Since time of day is an important factor for traffic flow, we decided to use it for our decision model. Also, while analyzing the data we observed the average velocity drops suddenly when the traffic is congested (occupancy ratio goes higher) indicating the vehicles come to a stopping point, or the traffic is now moving very slowly. With that, the number of cars passing (every 2 minutes) goes down. In order to quantify this phenomena, we decided to include the “road capacity” factor in our model. Road capacity basically provides the maximum number of vehicles that can pass within a given time period. We used the following relational formula for road capacity:

\begin{equation}
\small{Road Capacity \approx \# passing~cars (every~2~minutes) * avg.~velocity}
\end{equation}

We also observed that the average velocity is topped around 120-130 km/h regardless of the time, number of cars passing and road occupancy. As a result the road capacity is bounded. This information can be very useful for the prediction of traffic jam. 

\subsection{Feature Extraction}

The data from the sensors and the event logs are combined together to extract the most appropriate features for the accident detection model. After careful analysis, we observed that the differences of velocities, number of passing cars and capacity usage between consecutive readings are more important than the actual values. As a result, we decided to use the differential feature values. The selected features are as follows:

\begin{itemize}
    \item Average Velocity difference between reading at time T and T+1.
    \item Average Occupancy difference between reading at time T and T+1.
    \item Average Capacity usage difference between reading at time T and T+1.
    \item Weekday/Weekend (1 for Weekday, 0 for Weekend)
    \item Rush Hour (1 full rush hour, 0 late night traffic, other values depend on then time of day)
    \item Occurence of Accident/event (0 Normal traffic, 1 Accident/event)
\end{itemize}

The first 5 features are used as model inputs, the last one is the output feature that is to be classified by the selected model.

\section{Accident Detection Model}

Three different computational intelligence models are analyzed for accident detection. The selected models are nearest neighbor, regression tree, and feedforward neural networks. 

Nearest neighbor is one of the simplest classifiers that has been widely used for pattern recognition/classification problems~\cite{russell2010}. There is actually no learning involved in the algorithm. The model uses a number of representative data points for reference purposes, then it labels the new incoming data to the class of the nearest reference point based on their proximity in the input vector space. Euclidean distance is mostly used as the metric for this model. 

Regression tree is another simple classifier that uses the entropy and/or class variance  information for machine learning problems~\cite{russell2010}. The model consecutively selects the feature that best separates the classes based on minimum entropy principle. After each step, the remaining data within the branch is recursively processed through the same algorithm. As a result, a tree is formed. When a new data is introduced, the tree is traversed by answering a bunch of Yes/No questions at each step, until the corresponding class is found.

Feedforward neural networks are commonly used in complex nonlinear data mapping, function approximation, machine learning, and classification problems~\cite{russell2010}. There are many different implementations of neural networks in various different application areas. The basic principle behind such a model is to find the optimum weight values between the neurons that best match the input-output relations. Backpropagation learning algorithm is by far the most widely used method in network weight update process by error minimization at the output through iterations.

In this study, these three models are used for accident detection problem. There are 72 accident/event reports available for the year 2015 for the aforementioned highway section. A total of 276,354 data points are used for the model. Since the data was highly imbalanced, we used 130 data points for training for nearest neighbor and regression tree (of which 58 of them belong to the accidents). For the feedforward neural network model we used 100 data points for training, of which 42 of them belong to the accidents, 30 data points for cross validation, of which 16 of them belong to the accidents. The remaining data was used for testing purposes, of which there were 14 accidents and 276,224 non-accidents.  The confusion matrices of the test data for each model are tabulated in the following tables. Since the number of false alarms (model predicts an accident, but  actually there is none) is a lot, different loss threshold values are selected to find a trade-off between the false alarms and actual accident cases. The Tables~\ref{tbl:nnm}-\ref{tbl:ffnnm5} represent these different cases. Table~\ref{tbl:performance} provides the performance summary of these different selections. In the table TPR=TP/(TP+FN), TNR=TN/(TN+FP), PPV=TP/(TP+FN), NPV=TN/(TN+FN), and Accuracy=(TP+TN)/(TP+TN+FP+FN) where TP is true positives (accidents predicted as accidents, FN is false negatives (accidents predicted as non-accidents), FP is false positives (non-accidents predicted as accidents), and TN is true negatives (non-accidents predicted as non-accidents). 

\begin{table}[]
\centering
\caption{Nearest Neighbor Model Confusion Matrix}
\label{tbl:nnm}
\begin{tabular}{|l|l|l|l|}
\hline
\multicolumn{2}{|l|}{\multirow{2}{*}{Loss = 0 (Equal bias)}} & \multicolumn{2}{c|}{Predicted} \\ \cline{3-4} 
\multicolumn{2}{|l|}{}                                       & Accident     & Non-Accident    \\ \hline
\multicolumn{1}{|c|}{\multirow{2}{*}{Actual}} & Accident     & 13           & 1               \\ \cline{2-4} 
\multicolumn{1}{|c|}{}                        & Non-Accident & 13,468       & 262,742         \\ \hline
\end{tabular}
\end{table}

\begin{table}[]
\centering
\caption{Regression Tree Model Confusion Matrix}
\label{tbl:rtm}
\begin{tabular}{|l|l|l|l|}
\hline
\multicolumn{2}{|l|}{\multirow{2}{*}{Loss = 0 (Equal bias)}} & \multicolumn{2}{c|}{Predicted} \\ \cline{3-4} 
\multicolumn{2}{|l|}{}                                       & Accident     & Non-Accident    \\ \hline
\multicolumn{1}{|c|}{\multirow{2}{*}{Actual}} & Accident     & 13           & 1               \\ \cline{2-4} 
\multicolumn{1}{|c|}{}                        & Non-Accident & 8,885       & 267,325         \\ \hline
\end{tabular}
\end{table}

\begin{table}[]
\centering
\caption{Regression Tree Model Confusion Matrix}
\label{tbl:rtm2}
\begin{tabular}{|l|l|l|l|}
\hline
\multicolumn{2}{|l|}{\multirow{2}{*}{Loss = 0.5 (less alarms)}} & \multicolumn{2}{c|}{Predicted} \\ \cline{3-4} 
\multicolumn{2}{|l|}{}                                       & Accident     & Non-Accident    \\ \hline
\multicolumn{1}{|c|}{\multirow{2}{*}{Actual}} & Accident     & 12           & 2               \\ \cline{2-4} 
\multicolumn{1}{|c|}{}                        & Non-Accident & 6,652       & 269,558         \\ \hline
\end{tabular}
\end{table}

\begin{table}[]
\centering
\caption{Feedforward Neural Network  Model Confusion Matrix}
\label{tbl:ffnnm}
\begin{tabular}{|l|l|l|l|}
\hline
\multicolumn{2}{|l|}{\multirow{2}{*}{Loss = 0 (equal bias) 20 hidden neurons }} & \multicolumn{2}{c|}{Predicted} \\ \cline{3-4} 
\multicolumn{2}{|l|}{}                                       & Accident     & Non-Accident    \\ \hline
\multicolumn{1}{|c|}{\multirow{2}{*}{Actual}} & Accident     & 11           & 3               \\ \cline{2-4} 
\multicolumn{1}{|c|}{}                        & Non-Accident & 4,470       & 271,440         \\ \hline
\end{tabular}
\end{table}

\begin{table}[]
\centering
\caption{Feedforward Neural Network  Model Confusion Matrix}
\label{tbl:ffnnm2}
\begin{tabular}{|l|l|l|l|}
\hline
\multicolumn{2}{|l|}{\multirow{2}{*}{Loss = 0.5 (less alarms) 5 hidden neurons }} & \multicolumn{2}{c|}{Predicted} \\ \cline{3-4} 
\multicolumn{2}{|l|}{}                                       & Accident     & Non-Accident    \\ \hline
\multicolumn{1}{|c|}{\multirow{2}{*}{Actual}} & Accident     & 11           & 3               \\ \cline{2-4} 
\multicolumn{1}{|c|}{}                        & Non-Accident & 3,988       & 272,222         \\ \hline
\end{tabular}
\end{table}

\begin{table}[]
\centering
\caption{Feedforward Neural Network  Model Confusion Matrix}
\label{tbl:ffnnm3}
\begin{tabular}{|l|l|l|l|}
\hline
\multicolumn{2}{|l|}{\multirow{2}{*}{Loss = 0 (equal bias) 10 hidden neurons }} & \multicolumn{2}{c|}{Predicted} \\ \cline{3-4} 
\multicolumn{2}{|l|}{}                                       & Accident     & Non-Accident    \\ \hline
\multicolumn{1}{|c|}{\multirow{2}{*}{Actual}} & Accident     & 12           & 2               \\ \cline{2-4} 
\multicolumn{1}{|c|}{}                        & Non-Accident & 4,514       & 271,696         \\ \hline
\end{tabular}
\end{table}

\begin{table}[]
\centering
\caption{Feedforward Neural Network  Model Confusion Matrix}
\label{tbl:ffnnm4}
\begin{tabular}{|l|l|l|l|}
\hline
\multicolumn{2}{|l|}{\multirow{2}{*}{Loss = 0.5 (less alarms) 20 hidden neurons }} & \multicolumn{2}{c|}{Predicted} \\ \cline{3-4} 
\multicolumn{2}{|l|}{}                                       & Accident     & Non-Accident    \\ \hline
\multicolumn{1}{|c|}{\multirow{2}{*}{Actual}} & Accident     & 7          & 7               \\ \cline{2-4} 
\multicolumn{1}{|c|}{}                        & Non-Accident & 1,363       & 274,847         \\ \hline
\end{tabular}
\end{table}

\begin{table}[]
\centering
\caption{Feedforward Neural Network  Model Confusion Matrix}
\label{tbl:ffnnm5}
\begin{tabular}{|l|l|l|l|}
\hline
\multicolumn{2}{|l|}{\multirow{2}{*}{Loss = 0.94 (less alarms) 10 hidden neurons }} & \multicolumn{2}{c|}{Predicted} \\ \cline{3-4} 
\multicolumn{2}{|l|}{}                                       & Accident     & Non-Accident    \\ \hline
\multicolumn{1}{|c|}{\multirow{2}{*}{Actual}} & Accident     & 6           & 8               \\ \cline{2-4} 
\multicolumn{1}{|c|}{}                        & Non-Accident & 580       & 275,630         \\ \hline
\end{tabular}
\end{table}

\begin{table}[]
\centering
\caption{Performance comparison of different models}
\label{tbl:performance}
\begin{tabular}{|l|p{1cm}|p{1cm}|p{1cm}|p{1cm}|p{1cm}|}
\hline
Model &	Recall-True Positive Rate (TPR) (\%)	& Recall-True Negative Rate (TNR)(\%) &	Precision – Positive Predictive Value (PPV) (\%) & Precision – Negative Predictive Value  (NPV) (\%) & Accuracy (\%) \\
\hline
Table 1	& 92.86	& 95.12	& 0.09	& ≅ 100	& 95.12 \\
Table 2	& 92.86	& 96.78	& 0.15	& ≅ 100	& 96.78 \\
Table 3	& 85.71	& 97.59	& 0.18	& ≅ 100	& 97.59 \\
Table 4	& 78.57	& 98.27	& 0.23	& ≅ 100	& 98.27 \\
Table 5	& 78.57	& 98.56	& 0.28	& ≅ 100	& 98.56 \\
Table 6	& 85.71	& 98.37	& 0.27	& ≅ 100	& 98.37 \\
Table 7	& 50.00	& 99.51	& 0.51	& ≅ 100	& 99.50 \\
Table 8	& 42.86	& 99.79	& 1.02	& ≅ 100	& 99.79 \\
\hline
\end{tabular}
\end{table}

The results indicate that all models are very good in catching accidents, however the number of false positives are considerably high. When we increase the bias towards reducing these false alarms, the positive recall value starts going down, that is the trade-off we need to deal with. For example the last model presented in Table~\ref{tbl:ffnnm5}, has only 580 false alarms overall, however, with that setting, more than half of the accidents were not caught by the model.

New separate features may be included in the system that might improve the overall performance. Some delay can be introduced, so the sensor readings some time after the occurrence of the accidents can  be presented to the system, it might help in prediction performance, but it also results in the more delayed response time for the officials. More analyses need to be performed to achieve the best possible outcome.

One other problem is the involvements of the human factor in such a problem. We have realized, and told by the experts, that some minor problems on the road (some slippery point on the highway that causes the vehicles to slow down, an unreported minor accident, a handicapped car, unreported road work, suddenly changed weather conditions, etc.) might cause abnormal sensor readings, even sometimes the sensor readings themselves might have problems. According to the domain experts, these problems exist in their systems and some of the false positive data can be due to these cases. It is very difficult to come up with a clean data collection system; even if we eliminate the outliers totally, we will still have a lot of unconventional data points that are different than normal traffic flow. Engineers and data scientists will be constantly working on these type of issues in order to have better classifier models. Meanwhile, our models present preliminary results, but they can still be useful in the sense that, even though there are a lot of false alarms, just checking the status does not cost anything, however if there is actually an accident, it can assist a lot of people much faster. Just an example, Model 8 catches almost half of the accidents, and by average it gives between 1.5 false alarms per day. That might be a satisfactory system that the traffic department can tolerate. 

\section{Conclusion}

In this study, we implemented an automated accident detection system based on computational intelligence techniques. The data belongs to 2015 Istanbul highway sensor readings and traffic database. The data is processed and consolidated using big data processing techniques. Several different computational intelligence models are adapted and tested. Even though the number of false alarms is considerably high, the overall accuracy of the models are mostly over 99\%. This can provide early response to the accidents and save lives and valuable time/resources. The preliminary results indicate that it might be possible to use such a system in real-time at the Traffic Departments. For future work, it might be possible to add other independent features such as meteorological parameters, road topology information,relative location and/or condition of the road (direction of the sun, nearby buildings, road signs, service roads, inclination, existence of emergency lanes, etc.). Also more analyses need to be performed on the data itself, domain experts can be involved in the careful elimination of outlier points. Other machine algorithms can be adapted. The model can be represented as a time series problem, time warping can be used, recurrent neural network models can be implemented. Finally, the results from a combination of different approaches can be consolidated to have better prediction performance.

% if have a single appendix:
%\appendix[Proof of the Zonklar Equations]
% or
%\appendix  % for no appendix heading
% do not use \section anymore after \appendix, only \section*
% is possibly needed

% use appendices with more than one appendix
% then use \section to start each appendix
% you must declare a \section before using any
% \subsection or using \label (\appendices by itself
% starts a section numbered zero.)
%

% use section* for acknowledgement
%\section*{Acknowledgment}

%The authors would like to thank...

% Can use something like this to put references on a page
% by themselves when using endfloat and the captionsoff option.
\ifCLASSOPTIONcaptionsoff
  \newpage
\fi

% trigger a \newpage just before the given reference
% number - used to balance the columns on the last page
% adjust value as needed - may need to be readjusted if
% the document is modified later
%\IEEEtriggeratref{8}
% The "triggered" command can be changed if desired:
%\IEEEtriggercmd{\enlargethispage{-5in}}

% references section

% can use a bibliography generated by BibTeX as a .bbl file
% BibTeX documentation can be easily obtained at:
% http://www.ctan.org/tex-archive/biblio/bibtex/contrib/doc/
% The IEEEtran BibTeX style support page is at:
% http://www.michaelshell.org/tex/ieeetran/bibtex/
%\bibliographystyle{IEEEtran}
% argument is your BibTeX string definitions and bibliography database(s)
%\bibliography{IEEEabrv,../bib/paper}
%
% <OR> manually copy in the resultant .bbl file
% set second argument of \begin to the number of references
% (used to reserve space for the reference number labels box)

\printbibliography
%\nocite{*}

%\begin{thebibliography}{1}

%\bibitem{IEEEhowto:kopka}
%H.~Kopka and P.~W. Daly, \emph{A Guide to \LaTeX}, 3rd~ed.\hskip 1em plus
%  0.5em minus 0.4em\relax Harlow, England: Addison-Wesley, 1999.

%\end{thebibliography}

% biography section
% 
% If you have an EPS/PDF photo (graphicx package needed) extra braces are
% needed around the contents of the optional argument to biography to prevent
% the LaTeX parser from getting confused when it sees the complicated
% \includegraphics command within an optional argument. (You could create
% your own custom macro containing the \includegraphics command to make things
% simpler here.)
%\begin{biography}[{\includegraphics[width=1in,height=1.25in,clip,keepaspectratio]{mshell}}]{Michael Shell}
% or if you just want to reserve a space for a photo:

\begin{IEEEbiography}[{\includegraphics[width=1in,height=1.25in,clip,keepaspectratio]{picture}}]{John Doe}
\blindtext
\end{IEEEbiography}

% You can push biographies down or up by placing
% a \vfill before or after them. The appropriate
% use of \vfill depends on what kind of text is
% on the last page and whether or not the columns
% are being equalized.

%\vfill

% Can be used to pull up biographies so that the bottom of the last one
% is flush with the other column.
%\enlargethispage{-5in}

% that's all folks
\end{document}